\documentclass[10pt, conference, letterpaper]{IEEEtran}
\IEEEoverridecommandlockouts
\usepackage{cite}
\usepackage{amsmath,amssymb,amsfonts}
\usepackage{graphicx}
\usepackage{textcomp}
\usepackage{xcolor}
\usepackage[table]{xcolor}
\def\BibTeX{{\rm B\kern-.05em{\sc i\kern-.025em b}\kern-.08em
    T\kern-.1667em\lower.7ex\hbox{E}\kern-.125emX}}
\usepackage{siunitx}
\usepackage{booktabs}
\usepackage{multirow}
\usepackage{tabularx}  
\usepackage{array}     
\usepackage[ruled,vlined]{algorithm2e}
\usepackage{float} 
\usepackage{makecell}
\usepackage{url}
\usepackage{threeparttable}
\usepackage{caption}
\usepackage{pifont}
\usepackage{enumitem}
\usepackage{hyperref}
\hypersetup{
    breaklinks = true,        
    colorlinks = true,        
    linkcolor = red,         
    citecolor = red,         
    urlcolor = red,          
    hidelinks = true, 
}

\def\system{FitOne}
\newcommand{\eg}{{e.g.}}

\begin{document}

\title{Enhancing Fitness Intelligence through Domain-Specific LLM Post-Training}

\author{\IEEEauthorblockN{1\textsuperscript{st} Xingtao Zhao$^*$ \thanks{$^*$Corresponding author}}
\IEEEauthorblockA{\textit{School of Cyber Science and Technology} \\
\textit{Beihang University}\\
Beijing, China \\
xtaozhao@buaa.edu.cn}
\and
\IEEEauthorblockN{2\textsuperscript{nd} Tian Yang}
\IEEEauthorblockA{\textit{School of Information} \\
\textit{Renmin University of China}\\
Beijing, China\\
yangt@ruc.edu.cn}
\and
\IEEEauthorblockN{3\textsuperscript{nd} Han Jiang}
\IEEEauthorblockA{\textit{School of Cyber Science and Technology} \\
\textit{Beihang University}\\
Beijing, China\\
hanjiang950@gmail.com}
}

\maketitle

\begin{abstract}
Scientific Fitness Coaching (SFC) is typically delivered by human professionals, making it costly and inaccessible to many. While recent advances in Large Language Models (LLMs) show considerable promise for more inclusive fitness coaching, directly deploying prevailing general-purpose LLMs in SFC reveals critical limitations. These models often lack sufficient domain-specific knowledge integration, leading to weak performance on complex SFC scenarios.
In this paper, we introduce \textbf{\system{}}, a series of fitness LLMs (with 8B and 32B parameters) designed to improve reliability and domain specialization for SFC applications. Built upon the Qwen3 foundation models, \system{} is developed through a three-stage post-training pipeline consisting of continual pre-training, supervised fine-tuning, and reinforcement learning, using large-scale, high-quality datasets derived from rigorous knowledge engineering. We conduct comprehensive evaluations of \system{} on professional fitness certification exams, including ACSM-EP and NSCA-CSCS, as well as general capabilities such as knowledge reasoning and instruction following. Experimental results show that, while retaining strong general capabilities, \system{}-8B/32B achieves average improvements of up to 10.09\%/9.29\% and 12.73\%/7.01\% on the ACSM-EP and NSCA-CSCS exams, respectively, compared with the Qwen3 base models. Furthermore, in-depth ablation studies confirm the necessity of each training stage, highlighting the pipeline's effectiveness in balancing domain expertise enhancement with general ability retention.
We believe this research advances LLM systems toward more reliable fitness intelligence and will inspire future research on developing domain-specific LLMs.

\end{abstract}


\section{Introduction}
Driven by growing health awareness, fitness has evolved from a simple exercise routine into a health concept deeply integrated into people's daily lives~\cite{kramer2020overview,a20242025}. An increasing number of people rely on scientific fitness to improve their body aesthetics, physical function, and overall well-being including cardiovascular, metabolic, immune and mental health~\cite{mehta2025optimizing, yermolenko2024fitness}. However, achieving these goals remains challenging for the general public, as it requires specialized knowledge spanning exercise physiology, sports medicine, and nutrition~\cite{de2011more, xu2022understanding}. The gold standard for Scientific Fitness Coaching (SFC) is one-on-one guidance from credentialed professionals~\cite{oi2024qualified}, such as those holding ACSM-EP (American College of Sports Medicine Certified Exercise Physiologist)~\cite{acsmepcertifications} or NSCA-CSCS (National Strength and Conditioning Association Certified Strength and Conditioning Specialist)~\cite{nscacscscertifications} certifications. Despite their effectiveness, such human-led coaching services are often expensive, inaccessible to many, and difficult to scale~\cite{pelletier2020implementation}.

Recently, Large Language Models (LLMs) have achieved rapid performance improvements~\cite{chang2024survey,deepseekv32,qwen35,bytedanceseed2,gemini31flash,claudehaiku45,openaigpt54}, creating new opportunities to address the aforementioned challenges in SFC. However, directly applying general-purpose LLMs to this domain reveals critical limitations~\cite{info:doi/10.2196/59309,nazi2024large}. Unlike conventional natural language generation tasks, SFC requires not only the integration of interdisciplinary knowledge but also verifiable, step-by-step reasoning capabilities due to its safety-sensitive nature. In practice, professional fitness coaches are responsible for dynamically tailoring training plans based on users' physiological conditions and exercise goals, adjusting exercise programs when health constraints arise or recovery status changes, and designing targeted nutritional strategies to support post-training recovery~\cite{acsmepcertifications,nscacscscertifications,nsca2021essentials}. 
General reasoning LLMs often struggle with such complex, multi-faceted tasks because they lack deep integration of domain-specific knowledge. Although several studies have explored adapting LLMs to fitness applications~\cite{shin2025planfitting,NarratingFitness2024,khasentino2025personal,merrill2024transforming,heydari2025anatomypersonalhealthagent}, these solutions typically focus on narrow sub-fields and therefore struggle to meet the comprehensive demands of real-world SFC.

To address the challenges faced by general-purpose LLMs in SFC applications, we introduce \textbf{\system{}}, a family of domain-optimized fitness LLMs built upon the Qwen3~\cite{qwen3} foundation models. \system{} is developed through a domain-specialized, three-stage post-training pipeline utilizing large-scale, high-quality datasets constructed through rigorous knowledge engineering. The pipeline comprises cntinual pre-training (CPT), supervised fine-tuning (SFT), and reinforcement learning (RL). In the CPT stage, the model acquires comprehensive SFC knowledge from a carefully structured large-scale domain corpus. Building on this foundation, we enhance the model's step-by-step reasoning capabilities in the SFT stage through training on a high-quality fitness reasoning dataset. Finally, in the RL stage, we further consolidate prior gains and stabilize model behavior via Decoupled Clip and Dynamic sAmpling Policy Optimization (DAPO)~\cite{yu2025dapoopensourcellmreinforcement} algorithm, maximizing its practical utility in real-world deployments.

Extensive experiments demonstrate that \system{} not only maintains competitive results on general capability benchmarks, but also achieves superior performance across professional fitness certification exams, significantly outperforming prevailing closed- or open-source LLMs. Further ablation studies verify the necessity of each training stage and validate the effectiveness of our domain-specialized post-training pipeline in enhancing domain expertise without catastrophic forgetting. Our contributions can be summarized as follows:
\begin{itemize}
    \item We introduce \system{}, a domain-specific model series (8B and 32B parameters) built upon the Qwen3 foundation models, which effectively improves the reliability and domain specialization of LLMs in SFC applications.
    \item We propose a three-stage post-training pipeline that can effectively enhance the model’s performance in SFC scenarios while maintaining strong general capabilities. 
    \item Extensive experiments and ablation studies demonstrate the effectiveness of \system{} and our domain-specialized training methodology.
\end{itemize}

\section{\system{} Models}
\begin{figure*}
    \centering
    \includegraphics[width=0.9\linewidth]{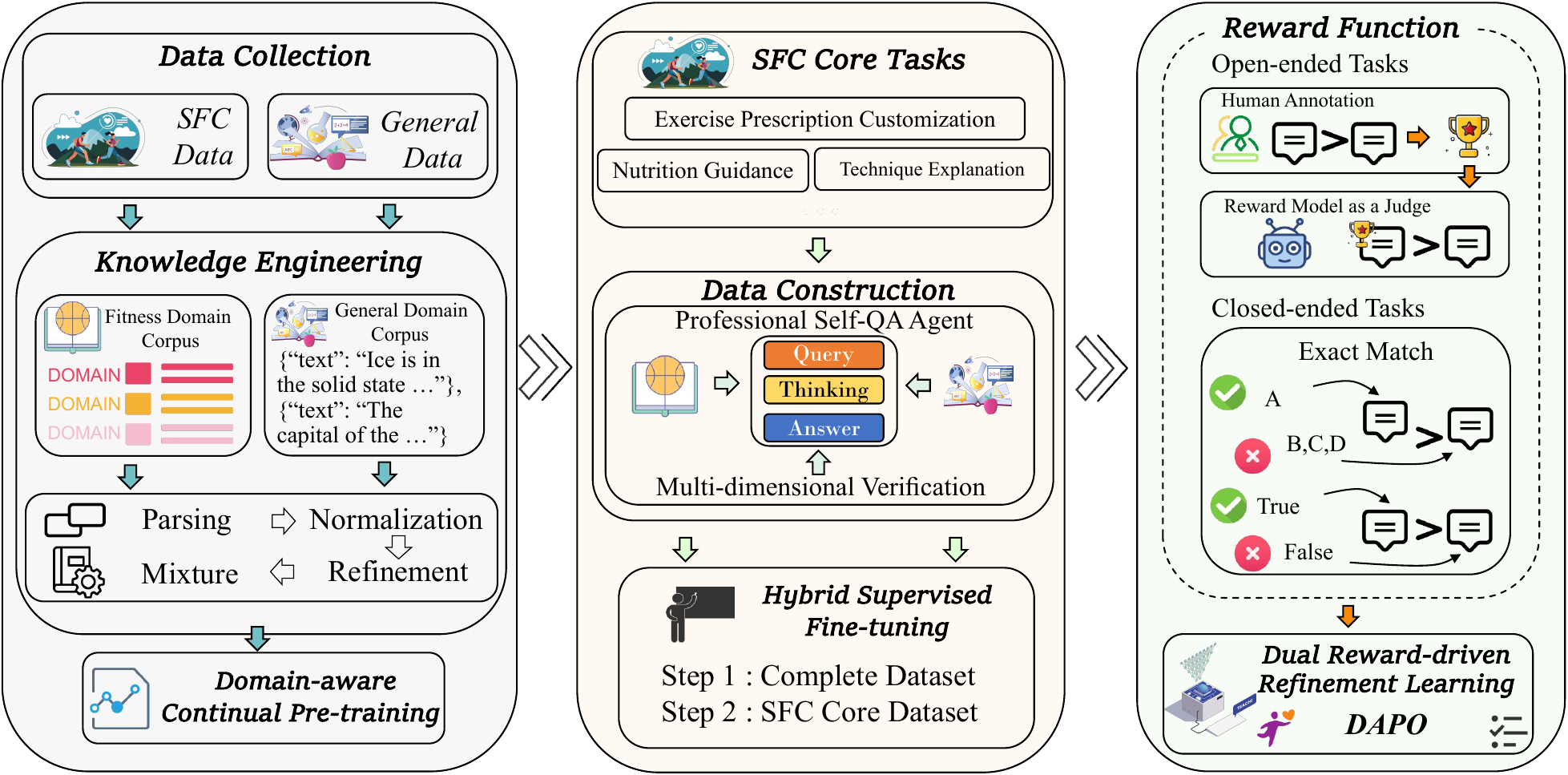}
    \caption{Overview of our training pipeline.}
    \label{fig:pipeline}
\end{figure*}

As shown in Figure \ref{fig:pipeline}, the post-training pipeline of \system{} consists of three sequential stages, each building on the preceding one to gradually adapt the base model to the SFC domain while preserving its general capabilities. First, in Section~\ref{sec:cpt}, we conduct continual pre-training to enhance the model’s grasp of foundational scientific fitness knowledge. Subsequently, in Section~\ref{sec:sft}, we further optimize the model’s verifiable reasoning capabilities through supervised fine-tuning using a high-quality chain-of-thought (CoT) dataset. Finally, in Section~\ref{sec:rl}, we apply DAPO-based reinforcement learning to consolidate the model’s previously acquired professional competencies and align its outputs with practical SFC requirements, ultimately yielding \system{} with superior real-world performance in SFC scenarios.

\subsection{Continual Pre-training}\label{sec:cpt}
To enhance the base model's fundamental domain knowledge, we conduct continual pre-training through three sub-stages: data collection, knowledge engineering, and domain-aware continual pre-training.

\subsubsection{Data Collection}
We curate a large-scale corpus from both the SFC domain and the general domain. To ensure comprehensive coverage of the fitness domain, under the guidance of domain experts, we categorize SFC into eight key sub-domains (Table~\ref{tab:SFC_category}), each corresponding to a core capability essential for SFC. Specifically, we collect the SFC corpus from authoritative sources to ensure scientific rigor, including well-established exercise guidelines(\eg{}, ACSM's Guidelines for Exercise Testing and Prescription~\cite{ACSMguidelines202512th}), peer-reviewed scientific literature, and professional university textbooks. This establishes a solid data foundation for adapting \system{} to a wide range of real-world SFC applications. For the general-domain data, we integrate open-source datasets that have been widely validated by the community~\cite{fan2025megasciencepushingfrontiersposttraining,weber2024redpajama,penedo2024fineweb} to prevent catastrophic forgetting during continual pre-training.

\begin{table*}[htbp]
\centering
\caption{Overview of SFC domains and their corresponding capabilities.}
\label{tab:SFC_category}
\small
\resizebox{0.9\textwidth}{!}{
\begin{tabular}{ll}
\hline
\textbf{Domain} & \textbf{Capability} \\
\hline
Weight Loss & Body weight management and body composition optimization \\
Biochemistry & Elucidating how physical activity impacts internal physiological processes \\
Sports Medicine & Sports injury prevention and rehabilitation \\
Sports Nutrition & Evidence-based personalized exercise nutrition guidance \\
Exercise Physiology & Analysis of body system adaptive changes to exercise \\
Strength and Endurance & Muscle strength and physical endurance enhancement \\
Exercise Methods and Techniques & Optimizing physical performance through diversified exercise methods \\
Exercise Prescription Customization & Generating exercise plans based on Frequency, Intensity, Time, and Type principles~\cite{ACSMguidelines202512th}\\
\hline
\end{tabular}
}
\end{table*}

\subsubsection{Knowledge Engineering}  
Then, the raw data undergoes comprehensive knowledge engineering to ensure authenticity and domain alignment through a systematic four-step preprocessing pipeline:  
\begin{enumerate}  
\item \textbf{Data Parsing:} Convert raw documents with diverse formats into the unified, machine-readable Markdown format using Optical Character Recognition (OCR) and NLP techniques, removing irrelevant layout elements like page numbers and headers.
\item \textbf{Data Normalization:} Standardize heterogeneous data formats and reconstruct data structures to achieve uniform fitness data representation, thereby facilitating downstream semantic understanding.  
\item \textbf{Data Refinement:} Remove outdated, noisy, and duplicate content from the dataset to generate a high-fidelity, refined corpus.  
\item \textbf{Data Mixture:} Apply the RegMix method~\cite{liuregmix} to identify an optimal data mixture distribution and filter out unnecessary data.  
\end{enumerate}  
Through this preprocessing process, we construct a high-quality training dataset of 30B tokens. The final refined dataset $\mathcal{K}$ comprises high-quality, structured fitness knowledge units that have undergone rigorous validation:  
\begin{equation}  
    \mathcal{K} = \mathcal{K}_{\text{SFC}} \cup \mathcal{K}_{\text{GEN}}, \mathcal{K} = \{k_1, k_2, \ldots, k_n\}   
\end{equation}  
where $\mathcal{K}_{\text{SFC}}$ and $\mathcal{K}_{\text{GEN}}$ denote the SFC-domain and general-domain subsets, respectively, each consisting of refined knowledge units $k_i$. 

\subsubsection{Domain-aware Continual Pre-training}
After data construction, we conduct continual pre-training on the complete dataset $\mathcal{K}$. Specifically, leveraging the strong language understanding capabilities inherent in the base Qwen3 model, \system{} is initialized from its original checkpoint and trained using the officially recommended configuration settings~\cite{qwen3}. Through this domain-aware continual pre-training process, we obtain a model that effectively captures SFC-specific linguistic patterns and domain knowledge while sustaining minimal degradation of general capabilities.

\subsection{Supervised Fine-Tuning}\label{sec:sft}
We further perform supervised fine-tuning through a reasoning-enhanced instruction dataset and a two-stage hybrid fine-tuning strategy. This fine-tuning process effectively enhances the model’s verifiable reasoning capabilities, which is essential for meeting the transparency and traceability requirements in safety-sensitive fitness applications.

\subsubsection{Data Construction}
We design a data synthesis pipeline that combines domain-specific knowledge-guided generation with multi-dimensional verification. The final instruction dataset comprises high-quality, verifiable \textit{(Query, Thinking, Answer)} triplets that are semantically aligned with the fitness domain.

\textbf{Domain-specific Knowledge-Guided Generation:}
According to Table~\ref{tab:SFC_category}, we define a structured label system of SFC tasks as $\mathcal{T} = \{t_1, t_2, \ldots, t_m\}$, where each label $t_j \in \mathcal{T}$ represents a distinct sub-domain category. For each $t_j$, a dedicated generation agent $A_{t_j}$ is instantiated. Given a label $t_j \in \mathcal{T}$ and a domain-specific knowledge snippet $k_i \in \mathcal{K}_{t_j} \subseteq \mathcal{K}_{\text{SFC}}$, the agent $A_{t_j}$ generates a reasoning triplet:
\begin{equation}
(q_i, t_i, a_i) = A_{t_j}(k_i, t_j; \theta_{A_{t_j}}),
\end{equation}
where $q_i$ is the question, $t_i$ is the step-by-step reasoning process (thinking), and $a_i$ is the final answer. This ensures the generated instructions are strictly grounded in professional fitness knowledge.

\textbf{Multi-dimensional Verification:}
To ensure data validity and reasoning reliability across all generated instances, we employ an independent third-party LLM to validate the correctness of answers and analyze the reasoning trajectory along seven key dimensions: internal consistency, term overlap rate, number of reasoning steps, logical coherence, content diversity, task-domain relevance, and alignment with task instructions. Based on these evaluations, we systematically filter the reasoning paths, retaining only high-quality instances, denoted as $\mathcal{D}_{\text{SFC}}$.

Furthermore, to maintain robust general capabilities and prevent catastrophic forgetting, we integrate open-source general instruction datasets~\cite{li2025infinityinstructscalinginstruction,guhaopenthoughts,yelimo}, denoted as $\mathcal{D}_{\text{GEN}}$. The combined dataset $\mathcal{D} = \mathcal{D}_{\text{SFC}} \cup \mathcal{D}_{\text{GEN}}$ is then utilized for SFT.

\subsubsection{Hybrid Supervised Fine-tuning}
We implement a two-stage hybrid SFT strategy to balance domain specialization and general capability retention. In the first stage, the model is trained on a broad mixture of the complete $\mathcal{D}_{\text{SFC}}$ and a large proportion of general data $\mathcal{D}_{\text{GEN}}$. This approach enables the model to learn diverse reasoning paths within the SFC domain while preserving its generalization capabilities. In the second stage, we increase the sampling weight of $\mathcal{D}_{\text{SFC}}$, fine-tuning the model to reinforce its performance on domain-critical tasks. Given an input question $Q$ and a target output sequence $Y = [T; A]$, the SFT objective is formally defined as: 
\begin{equation}
\mathcal{L}_{\text{SFT}} = - \mathbb{E}_{(Q, Y) \sim \mathcal{D}} \left[ \sum_{t=1}^{|Y|} \log \pi_\theta(y_t \mid Q, Y_{<t}) \right]
\end{equation}
where $\pi_\theta$ denotes the model policy and $y_t$ is the $t$-th token in the target sequence $Y$.

\subsection{Reinforcement Learning}\label{sec:rl}
To consolidate prior gains and align the model's behavior with real-world application requirements, we build upon the SFT model and conduct reinforcement learning using SFC-centric reward signals.

\subsubsection{Data Construction}
We construct the RL dataset, denoted as $\mathcal{D}_{\text{RL}}$, by combining subsets from both $\mathcal{D}_{\text{SFC}}$ and $\mathcal{D}_{\text{GEN}}$. To maximize RL efficiency, we upweight hard examples identified during SFT evaluations and rare but critical edge cases such as exercise scenarios with complex medical constraints. This targeted data mixture strategy ensures that the model focuses on its weakest reasoning areas rather than repeatedly training on mastered tasks.

\subsubsection{Reward Function Design}
Constructing a reliable reward function is the most critical component of our RL pipeline. While prior reinforcement learning algorithms~\cite{DeepSeekR1} rely heavily on rule-based rewards for domains with deterministic answers (e.g., mathematics and coding), SFC applications span both objective certification questions and open-ended exercise prescriptions. To accommodate this task diversity, we design a hybrid reward function for a given query $Q$, reference answer $A$, and model output $O$:

\textbf{1) Accuracy Reward.} 
This reward ensures factual correctness and aligns with human preferences, dynamically adapting based on task verifiability. For closed-ended tasks such as multiple-choice questions with determinate answers, we employ an Exact Match (EM) reward based on the ground truth $A$:
\begin{equation}
        \mathcal{R}_{\text{EM}}(O, A) = 
        \begin{cases}
        1, & \text{if } O \text{ exactly matches } A, \\
        0, & \text{otherwise}.
        \end{cases}
\end{equation}
For open-ended tasks such as personalized nutrition planning, where a single ground truth is infeasible, we train a reward model (RM) using domain-expert preference annotations to provide a continuous scalar reward.
\begin{equation}
        \mathcal{R}_{\text{RM}}(O,A) = \text{RM}(O,A)
\end{equation}

\textbf{2) Pattern Reward.} 
Additionally, given the instability of generative LLM output formats, we define a pattern-matching reward for tasks requiring strictly structured outputs. For instance, exercise prescriptions must explicitly cover the FITT-VP principles (i.e., Frequency, Intensity, Time, Type, Volume, Progression)~\cite{ACSMguidelines202512th}. We use predefined heuristics to verify structural compliance independently of semantic correctness:
\begin{equation}
    \mathcal{R}_{\text{Pattern}}(O,A) = 
    \begin{cases}
    1, & \text{if } O \text{ satisfies the predefined format}, \\
    0, & \text{otherwise}.
    \end{cases}
\end{equation}

Finally, the total reward $\mathcal{R}(O, A)$ combines both accuracy and pattern signals based on the specific task category $c$:
\begin{equation}
    \mathcal{R}(O, A) = \mathcal{R}_{\text{Acc}}(O, A, c) + \mathcal{R}_{\text{Pattern}}(O,A)
\end{equation}
where $\mathcal{R}_{\text{Acc}}$ applies either $\mathcal{R}_{\text{EM}}$ or $\mathcal{R}_{\text{RM}}$ depending on whether the task category $c$ is closed-ended or open-ended.

\subsubsection{DAPO-based Reinforcement Learning}
We initialize the RL policy $\pi_{\theta}$ from the prior stage to provide a strong starting base model, and then apply preference-based DAPO~\cite{yu2025dapoopensourcellmreinforcement}. Compared with the standard GRPO~\cite{DeepSeekR1}, DAPO's asymmetric clipping encourages broader exploration, and its dynamic sampling prevents gradient vanishing when all candidate outputs receive identical rewards, making it particularly effective for learning complex reasoning in SFC tasks. For a given instance $(Q, A) \sim \mathcal{D}_{\text{RL}}$, DAPO samples a group of $G$ candidate outputs $\{O_i\}_{i=1}^G$ using the reference policy $\pi_{\theta_{\mathrm{old}}}$. The policy $\pi_{\theta}$ is then updated by minimizing the following token-level loss:
\begin{align}
    \mathcal{L}_{\mathrm{DAPO}}(\theta)
    =& - \mathbb{E}_{(Q,A)\sim \mathcal{D}_{\text{RL}}, \{O_i\}_{i=1}^G \sim \pi_{\theta_{\mathrm{old}}}(\cdot \mid Q)} \nonumber\\
    &\Biggl[
    \frac{1}{\sum_{i=1}^G |O_i|}
    \sum_{i=1}^G \sum_{t=1}^{|O_i|}
    \min\Bigl(r_{i,t}(\theta)\,\hat A_i, \nonumber\\
    & \mathrm{clip}(r_{i,t}(\theta),\,1-\varepsilon_{\mathrm{low}},\,1+\varepsilon_{\mathrm{high}})\,\hat A_i\Bigr)
    \Biggr] \nonumber\\
    &\text{s.t.}\quad \mathrm{std}(\{\mathcal{R}_i\}_{i=1}^G) > 0
\end{align}
where $\varepsilon_{\mathrm{low}}$ and $\varepsilon_{\mathrm{high}}$ control the clipping range, and
\begin{equation}
    r_{i,t}(\theta) = \frac{\pi_\theta(O_{i,t} \mid Q, O_{i,<t})}{\pi_{\theta_{\mathrm{old}}}(O_{i,t} \mid Q, O_{i,<t})}
\end{equation}
\begin{equation}
    \hat A_i = \frac{\mathcal{R}_i - \mathrm{mean}(\{\mathcal{R}_i\}_{i=1}^G)}{\mathrm{std}(\{\mathcal{R}_i\}_{i=1}^G)}
\end{equation}

Compared to the previous stage, \system{} learns to consistently explore and exploit high-reward reasoning trajectories after reinforcement learning, ensuring stable and robust behavioral adaptation to both SFC-specific and general tasks.
\section{Experiments}
\begin{table*}[htbp]
\centering
\caption{
Results of \textbf{8B}-scale models.
\textbf{Bold} entries indicate the best model, while \underline{underlined} entries denote the second one. \emph{Total} represents the comprehensive score calculated based on the percentages of each domain, and \emph{Promotion} is the percentage improvement of FitOne over the Qwen3 foundation model. All results are the average values of five repeated experiments.
}
\resizebox{\textwidth}{!}{ 
\begin{tabular}{l|c|cccc|c|ccccccc|c}
\toprule
\multirow{2}{*}{\textbf{Models}} & 
\multicolumn{1}{c|}{\textbf{General-Bench}} & 
\multicolumn{5}{c|}{\textbf{ACSM-EP}} &
\multicolumn{8}{c}{\textbf{NSCA-CSCS}}\\
\cmidrule(lr){2-2}
\cmidrule(lr){3-7}
\cmidrule(lr){8-15}
& {\textbf{Avg.}} &
 {\textbf{HFS}} & 
  {\textbf{EC\&BM}} & 
 {\textbf{EPI}} & 
 {\textbf{RM\&PR}} &
 {\textbf{Total}}&
 {\textbf{ES}} & 
 {\textbf{SP}} & 
 {\textbf{Nutr}} & 
 {\textbf{PD}} &
 {\textbf{ET}} &
 {\textbf{PI}} &
 {\textbf{OA}} &
 {\textbf{Total}} \\
\midrule

Llama-3.1-8B \cite{llama3} & 52.41 & 45.12 & 46.99 & 37.19 & 44.21 & 42.26 & 51.80 & 66.16 & 56.20 & 53.87 & 63.63 & 89.31 & 51.51 & 60.37 \\
Ministral-3-8B \cite{ministral3} & 50.93 & 41.93 & 59.04 & 45.21 & 40.91 & 46.59 & 66.93 & 71.82 & 58.90 & 54.61 & 52.67 & 87.43 & 48.87 & 63.03 \\
InternLM3-8B \cite{internlm2} & 55.35 & 44.44 & 54.32 & 30.02 & 30.18 & 39.65 & 56.14 & 66.34 & 59.32 & 44.35 & 52.43 & 74.03 & 48.12 & 55.69 \\
GLM-4-9B-0414 \cite{glm2024} & 62.27 & 51.44 & 55.26 & \underline{48.83} & 42.87 & 50.56 & \underline{69.44} & 73.30 & 56.86 & 68.99 & 50.11 & 75.99 & 47.30 & 65.12 \\

\midrule

Qwen3-8B \cite{qwen3} & \underline{67.90} & \underline{52.64} & \underline{59.84} & 45.93 & \underline{55.36} & \underline{51.59} & 68.18 & \underline{74.83} & \underline{62.76} & \underline{69.03} & \underline{64.12} & \underline{89.98} & \underline{53.84} & \underline{69.64} \\
\rowcolor[HTML]{E6E6FA}
FitOne-8B \textbf{(Ours)} & \textbf{70.83} & \textbf{54.21} & \textbf{61.77} & \textbf{56.44} & \textbf{56.78} & \textbf{56.79} & \textbf{72.78} & \textbf{87.26} & \textbf{67.68} & \textbf{78.68} & \textbf{79.72} & \textbf{90.59} & \textbf{71.57} & \textbf{78.51} \\
Promotion & \textcolor{orange}{+4.32\%} & \textcolor{orange}{+2.98\%} & \textcolor{orange}{+3.23\%} & \textcolor{orange}{+22.88\%} & \textcolor{orange}{+2.57\%} & \textcolor{orange}{+10.09\%} & \textcolor{orange}{+6.75\%} & \textcolor{orange}{+16.61\%} & \textcolor{orange}{+7.84\%} & \textcolor{orange}{+13.98\%} & \textcolor{orange}{+24.33\%} & \textcolor{orange}{+0.68\%} & \textcolor{orange}{+32.93\%} & \textcolor{orange}{+12.73\%} \\
\bottomrule
\end{tabular}
}
\captionsetup{justification=raggedright,singlelinecheck=false,font=scriptsize}
\caption*{* \textbf{\href{https://acsm.org/wp-content/uploads/2024/12/ACSM-Certified-Exercise-Physiologist-Exam-Content-Outline.pdf}{ACSM-EP Performance Domains}:} 
\textbf{HFS (33\%)}: Health and Fitness Assessment \quad \textbf{EC\&BM (20\%)}: Exercise Counseling and Behavior Modification \\ 
\textbf{EPI (40\%)}: Exercise Prescription and Implementation \quad \textbf{RM\&PR (7\%)}: Risk Management and Professional Responsibilities \\
* \textbf{\href{https://www.nsca.com/cscs-exam-description/}{NSCA-CSCS Performance Domains}:} 
\textbf{ES (25\%)}:Exercise Science \quad \textbf{SP (11\%)}: Sport Psychology \quad \textbf{Nutr (6\%)}: Nutrition \quad \textbf{PD (23\%)}: Program Design \\ 
\textbf{ET (15\%)}: Exercise Technique  \quad  \textbf{PI (12\%)}: Program Implementation \quad \textbf{OA (8\%)}: Organization and Administration}
\label{table:8bmainresults}
\vspace{0.3em}
\end{table*}

\begin{table*}[htbp]
\centering
\caption{
Results of \textbf{32B}-scale models.
\textbf{Bold} entries indicate the best model, while \underline{underlined} entries denote the second one. \emph{Total} represents the comprehensive score calculated based on the percentages of each domain, and \emph{Promotion} is the percentage improvement of \system{} over the Qwen3 foundation model. All results are the average values of five repeated experiments.
}
\resizebox{\textwidth}{!}{ 
\begin{tabular}{l|c|cccc|c|ccccccc|c}
\toprule
\multirow{2}{*}{\textbf{Models}} & 
\multicolumn{1}{c|}{\textbf{General-Bench}} & 
\multicolumn{5}{c|}{\textbf{ACSM-EP}} &
\multicolumn{8}{c}{\textbf{NSCA-CSCS}}\\
\cmidrule(lr){2-2}
\cmidrule(lr){3-7}
\cmidrule(lr){8-15}
& {\textbf{Avg.}} &
 {\textbf{HFS}} & 
  {\textbf{EC\&BM}} & 
 {\textbf{EPI}} & 
 {\textbf{RM\&PR}} &
 {\textbf{Total}}&
 {\textbf{ES}} & 
 {\textbf{SP}} & 
 {\textbf{Nutr}} & 
 {\textbf{PD}} &
 {\textbf{ET}} &
 {\textbf{PI}} &
 {\textbf{OA}} &
 {\textbf{Total}} \\
\midrule
\multicolumn{15}{c}{\textit{10B$<$The Scale of Large Language Models$<$100B}} \\
\midrule
Gemma-3-27B \cite{gemma3technicalreport} 
& 75.71 & 64.72 & 69.43 & 64.29 & 66.20 & 65.59 & 76.61 & 82.38 & 77.41 & 73.24 & 76.41 & 94.95 & 71.02 & 78.24 \\
GPT-OSS-20B \cite{gptoss} 
& 74.76 & 62.74 & 70.58 & 65.41 & 62.88 & 65.39 & 70.18 & 82.89 & 79.15 & 79.37 & 75.78 & 93.36 & 78.52 & 78.52 \\
Qwen3-30B-A3B \cite{qwen3} 
& 72.76 & 62.60 & 66.67 & 60.37 & 62.37 & 62.51 & 71.41 & 81.91 & 74.11 & 73.35 & 76.93 & 92.44 & 73.13 & 76.66 \\
GLM-4-32B-0414 \cite{glm4} 
& 73.49 & 70.69 & 64.06 & 67.50 & 66.35 & 67.78 & 73.77 & 78.63 & 74.22 & 76.00 & 77.37 & 92.05 & 78.94 & 77.99 \\

\midrule
\multicolumn{15}{c}{\textit{The Scale of Large Language Models$>$100B}} \\
\midrule
Deepseek-V3.2 \cite{deepseekv32} 
& 78.02 & 71.22 & 73.88 & 73.07 & 75.05 & 72.76 & \textbf{78.75} & 81.54 & 84.29 & 79.16 & \underline{84.14} & 93.93 & 80.20 & 82.23 \\
Doubao-Seed-2-Lite \cite{bytedanceseed2} 
& 80.13 & 73.18 & 72.58 & 71.71 & 71.21 & 72.33 & 75.22 & 86.12 & 84.92 & \underline{80.78} & 82.04 & \textbf{99.40} & 80.71 & \underline{82.64} \\
Gemini-3.1-Flash \cite{gemini31flash} 
& 80.25 & 69.44 & 70.96 & 72.98 & 71.11 & 71.28 & 73.32 & 81.94 & \textbf{89.93} & 80.67 & 79.57 & 93.02 & 82.29 & 80.97 \\
Qwen3.5-Flash \cite{qwen35} 
& \underline{84.42} & 71.82 & \textbf{75.41} & \underline{77.40} & 75.89 & \underline{75.05} & 72.00 & 83.05 & \underline{88.96} & 78.97 & 81.15 & 91.40 & \underline{82.46} & 80.37 \\
GPT-5.4-Mini \cite{openaigpt54} 
& 80.72 & 69.39 & 73.85 & 70.89 & 75.45 & 71.31 & 75.54 & \underline{87.20} & 88.79 & 76.23 & 76.59 & 91.90 & 79.14 & 80.19 \\
Claude-Haiku-4.5 \cite{claudehaiku45} 
& \textbf{85.10} & \underline{73.78} & 69.68 & 76.18 & \underline{75.95} & 74.07 & 69.10 & 85.87 & 86.15 & 77.10 & 83.43 & 91.20 & 76.88 & 79.23 \\

\midrule
Qwen3-32B \cite{qwen3} 
& 72.67 & 70.15 & 68.54 & 70.44 & 68.85 & 69.85 & 74.52 & 84.04 & 79.39 & 74.56 & 76.76 & 91.19 & 74.48 & 78.20 \\
\rowcolor[HTML]{E6E6FA}
\system{}-32B \textbf{(Ours)} 
& 75.19 & \textbf{74.76} & \underline{75.10} & \textbf{78.06} & \textbf{77.58} & \textbf{76.35} & \underline{77.02} & \textbf{87.99} & 86.76 & \textbf{80.92} & \textbf{85.15} & \underline{95.87} & \textbf{83.19} & \textbf{83.68} \\
Promotion 
& \textcolor{orange}{+3.47\%} & \textcolor{orange}{+6.57\%} & \textcolor{orange}{+9.57\%} & \textcolor{orange}{+10.82\%} & \textcolor{orange}{+12.68\%} & \textcolor{orange}{+9.29\%} & \textcolor{orange}{+3.35\%} & \textcolor{orange}{+4.70\%} & \textcolor{orange}{+9.28\%} & \textcolor{orange}{+8.53\%} & \textcolor{orange}{+10.93\%} & \textcolor{orange}{+5.13\%} & \textcolor{orange}{+11.69\%} & \textcolor{orange}{+7.01\%} \\
\bottomrule
\end{tabular}
\label{table:32bmainresults}
}
\vspace{0.3em}
\end{table*}

\subsection{Implementation Details}\label{sec:imp_details}
During the CPT stage, we follow the standard Qwen3~\cite{qwen3} training process using a mixed corpus of general and SFC-specific data.
In the SFT stage, we train the models for one epoch in both steps. We use a maximum sequence length of 8,192 tokens with sequence packing, a global batch size of 128, and a linear warm-up ratio of 0.03. The learning rates are scaled based on model size: for the 8B model, we use \num{2e-5} in step one and \num{1e-5} in step two; for the 32B model, we use \num{1e-5} across both steps. Optimization is performed using AdamW~\cite{adamw} ($\beta_1=0.9$, $\beta_2=0.95$, $\epsilon=10^{-8}$).
In the RL stage, we train for 500 rollout steps. The maximum prompt and response lengths are set to 10,000 and 8,192 tokens, respectively, with an additional 4,096-token overlong buffer using a 1.0 penalty factor. We use a prompt batch size of 1,024 and sample 16 responses per prompt, resulting in a global batch size of 16,384. The mini-batch size covers 256 prompts, yielding 4 gradient updates per rollout. We set the asymmetric clipping parameters $\varepsilon_{\mathrm{low}}=0.2$ and $\varepsilon_{\mathrm{high}}=0.28$ and use AdamW optimizer with a constant learning rate of \num{1e-6}, a weight decay of 0.01, and a linear warmup over the first 20 rollout steps.

\subsection{Benchmarks}
We perform a comprehensive evaluation of \system{} and baselines in both the general and SFC domain capabilities using mainstream benchmarks.
To evaluate \system{}'s performance in SFC domain, we select professional fitness certification exams including \emph{ACSM-EP}~\cite{acsmepcertifications} and \emph{NSCA-CSCS}~\cite{nscacscscertifications}, covering eleven SFC performance domains.
For general capabilities evaluation, we systematically assess six capabilities, including knowledge reasoning (i.e., \emph{MMLU-Redux}~\cite{mmluredux}, \emph{CMMLU}~\cite{li2023cmmlu}, \emph{C-Eval}~\cite{ceval}, \emph{GPQA-Diamond}~\cite{rein2024gpqa}, \emph{BIG-Bench}~\cite{bbh}, and \emph{GAOKAO-Bench}~\cite{gaokaobench}), 
mathematical reasoning (i.e., \emph{MATH-500}~\cite{math}, \emph{AIME 2024}~\cite{aime} and \emph{AIME 2025}~\cite{aime}), 
code generation (i.e., \emph{HumanEval}~\cite{HumanEval}, \emph{MBPP}~\cite{mbpp}, and \emph{LiveCodeBench v5}~\cite{LiveCodeBench}), 
machine translation (i.e., \emph{WMT-24}~\cite{wmt24} and \emph{FLORES}~\cite{flores}), 
instruction following (i.e., \emph{IFEval}~\cite{ifeval}), 
and hallucination detection (i.e., \emph{HaluEval}~\cite{Halueval}).

\subsection{Main Results}
\begin{figure}[htbp]
    \centering
    \includegraphics[width=0.4\textwidth]{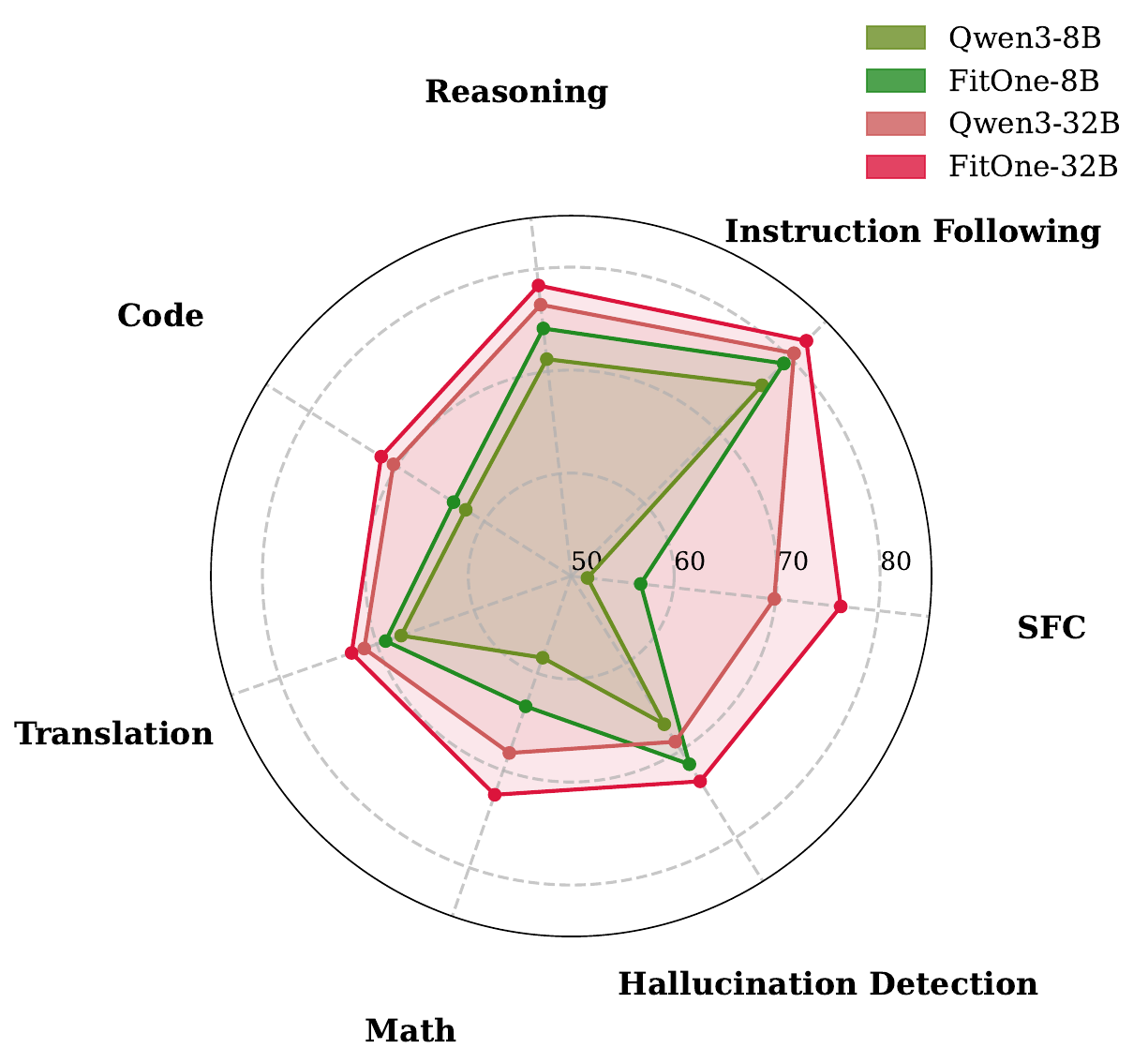}
    \caption{Radar chart of model capabilities under different task categories.}
    \label{fig:radar}
\end{figure}
Tables~\ref{table:8bmainresults} and \ref{table:32bmainresults} report the performance of \system{} against prevailing open- and closed-source models on both general competency assessment benchmarks and domain-specific professional fitness certification exams (ACSM-EP and NSCA-CSCS). Figure~\ref{fig:radar} further visualizes the comparative distribution of general capabilities compared to the Qwen3 base models. These results demonstrate that \system{} at all scales not only preserve strong general capabilities, often surpassing their base models, but also achieve substantial improvements in the SFC domain. 
Specifically, \system{}-8B scores 56.79 on the ACSM-EP and 78.51 on the NSCA-CSCS, outperforming the Qwen3-8B base model by 10.09\% and 12.73\%, respectively. This performance gain scales consistently to the 32B variant, aligning with established scaling laws~\cite{kaplan2020scalinglawsneurallanguage}. Notably, \system{}-32B matches even exceeds the SFC-domain performance of massive proprietary models with over 100B parameters, including DeepSeek-V3.2, Gemini-3.1-Flash, and GPT-5.4-Mini, with minor performance gaps observed in only a few specific sub-domains. These findings validate the effectiveness of our post-training pipeline, highlighting both its potential for further advances through continued scaling and \system{}'s promise for real-world SFC deployment.

\subsection{Ablation Study}
\begin{table}[t]
\centering
\caption{
Ablation study results of \system{}-8B.
}
\resizebox{0.5\textwidth}{!}{
\begin{tabular}{ccc|ccc}
\toprule
\textbf{CPT} & \textbf{SFT} & \textbf{RL} & 
\multicolumn{1}{c}{\textbf{General}} & 
\multicolumn{1}{c}{\textbf{ACSM-EP}} &
\multicolumn{1}{c}{\textbf{NSCA-CSCS}} 
 \\
\midrule
 & & & 67.90 & 51.59 & 69.64 \\
{$\checkmark$} & & & 66.24 & 53.82 & 72.15 \\
& {$\checkmark$} & & 69.45 & 52.91 & 73.80 \\
{$\checkmark$} & {$\checkmark$} & & 70.12 & 55.48 & 76.92 \\
& {$\checkmark$} & {$\checkmark$} & 69.80 & 54.10 & 75.33 \\
{$\checkmark$} & {$\checkmark$} & {$\checkmark$} & 70.83 & 56.79 & 78.51 \\
\bottomrule
\end{tabular}
}
\label{tab:ablation}
\end{table}
We evaluate the incremental impact and necessity of each stage in our post-training pipeline based on the 8B model, with results summarized in Table~\ref{tab:ablation}. 
First, the CPT stage establishes a strong domain foundation, improving performance to 53.82 on the ACSM-EP and 72.15 on the NSCA-CSCS. However, this domain-specific knowledge injection causes a slight degradation in general capabilities (dropping from 67.90 to 66.24 on General-Bench) due to domain shift. Subsequent SFT effectively resolves this issue by enhancing the model's reasoning capabilities, raising domain scores to 55.48 and 76.92 while significantly recovering and boosting the General-Bench score to 70.12. Finally, the RL stage further aligns the model with practical requirements, resulting in the peak scores across all metrics.
Additionally, removing the CPT stage from the pipeline consistently yields sub-optimal domain performance compared to their full-pipeline counterparts. For instance, the complete CPT+SFT+RL pipeline outperforms the SFT+RL configuration by 2.69 points on the ACSM-EP and 3.18 points on the NSCA-CSCS. This confirms that foundational knowledge injection during CPT is an indispensable prerequisite for maximizing the model's domain expertise, validating the necessity of our three-stage design.

\subsection{Task-specific SFT Comparison}
\begin{table}[t]
\centering
\caption{Performance comparison of task-specific fine-tuned on Qwen3-8B and \system{}-8B.}
\resizebox{0.5\textwidth}{!}{
\begin{tabular}{cccc}
\toprule
\textbf{Models} & \textbf{HFS} & \textbf{EC\&BM} & \textbf{EPI} \\
\midrule
Qwen3-8B (Fine-tuned)      & 53.50 & 60.90 & 50.20 \\
\system{}-8B (Zero-shot)&  54.21 \scriptsize{(+1.33\%)} &  61.77 \scriptsize{(+1.43\%)} & 56.44 \scriptsize{(+12.43\%)} \\
\system{}-8B (Fine-tuned)    & \textbf{56.80}\scriptsize{(+6.17\%)} & \textbf{64.30}\scriptsize{(+5.58\%)} & \textbf{60.50}\scriptsize{(+20.52\%)} \\
\midrule
\textbf{Models} & \textbf{ES} & \textbf{PD} & \textbf{ET} \\
\midrule
Qwen3-8B (Fine-tuned)      & 70.50 & 73.40 & 72.10 \\
\system{}-8B (Zero-shot)& 72.78 \scriptsize{(+3.23\%)} & 78.68 \scriptsize{(+7.19\%)} & 79.72 \scriptsize{(+10.57\%)} \\
\system{}-8B (Fine-tuned)    & \textbf{75.60}\scriptsize{(+7.23\%)} & \textbf{82.40}\scriptsize{(+12.26\%)} & \textbf{83.50}\scriptsize{(+15.81\%)} \\
\bottomrule
\end{tabular}
}
\label{tab:task_specific_sft}
\end{table}
To evaluate whether a domain-adapted base model serves as a better initialization point for downstream tasks compared to a general-purpose model, we compare three setups at the 8B scale across six core sub-domains from the professional exams: (1) {Qwen3-8B (Fine-tuned)}: task-specific SFT applied directly to the Qwen3 base model; (2) {\system{}-8B (Zero-shot)}: our domain model evaluated directly without further downstream tuning; and (3) {\system{}-8B (Fine-tuned)}: task-specific SFT applied to our domain-adapted model.

As shown in Table~\ref{tab:task_specific_sft}, \system{}-8B (Fine-tuned) outperforms Qwen3-8B (Fine-tuned) across all evaluated domains. For instance, in the Program Design (PD) and Exercise Technique (ET) domains, \system{}-8B (Fine-tuned) achieves 82.40 and 83.50, outperforming the fine-tuned Qwen3-8B by 12.26\% and 15.81\% percentages, respectively. This demonstrates that our domain-aligned post-training provides a stronger foundation for task-specific adaptation.
Meanwhile, even without any task-specific tuning, the zero-shot \system{}-8B surpasses the fine-tuned Qwen3-8B across all selected sub-domains. Overall, these findings indicate that domain-specific post-training not only instills robust zero-shot capabilities but also raises the performance ceiling for downstream fine-tuning, validating the necessity of developing specialized foundational models for the SFC domain.

\subsection{Generalization Across Different Model Scales}
\begin{table}[t]
\centering
\caption{Performance generalization across different model scales and architectures. We report the average performance for General-Bench, ACSM-EP and NSCA-CSCS certification exams.}
\label{table:model_scales}
\resizebox{0.48\textwidth}{!}{
\begin{tabular}{lccc}
\toprule
\textbf{Models} & \textbf{General-Bench} & \textbf{ACSM-EP} & \textbf{NSCA-CSCS} \\
\midrule
Qwen3-4B & 64.12 & 47.35 & 63.80 \\
\system{}-4B & 66.25 & 52.40 & 65.15 \\
\midrule
Qwen3-8B & 67.90 & 51.59 & 69.64 \\
\system{}-8B & 70.83 & 56.79 & 78.51 \\
\midrule
Qwen3-14B & 70.15 & 58.42 & 74.10 \\
\system{}-14B & 71.80 & 64.30 & 80.65 \\
\midrule
Qwen3-32B & 72.67 & 69.85 & 78.20 \\
\system{}-32B & 75.19 & 76.35 & 83.68 \\
\midrule
Qwen3-30B-A3B & 72.76 & 62.51 & 76.66 \\
\system{}-30B-A3B & 76.85 & 73.15 & 86.40 \\
\bottomrule
\end{tabular}
}
\end{table}
Table~\ref{table:model_scales} shows that our three-stage post-training pipeline generalizes effectively across base models of varying scales and architectures. The pipeline consistently improves performance across all benchmarks, proving its robustness regardless of model capacity. Furthermore, scaling up the base model increases \system{}’s absolute performance, indicating that larger capacities better absorb domain-specific knowledge.  
Notably, the mixture-of-experts (MoE) variant (Qwen3-30B-A3B) achieves the largest gains on SFC-related benchmarks, improving by 17.02\% on ACSM-EP and 12.71\% on NSCA-CSCS, while maintaining top-tier general capabilities.
We think this is because the decoupled expert routing mechanism of the MoE architecture provides an expansive representational space to internalize complex, interdisciplinary SFC domain knowledge. This architecture effectively mitigates gradient conflicts between general and domain-specific data distributions during post-training, preventing catastrophic forgetting and enabling robust domain adaptation.



\section{Conclusion}
In this paper, we introduce \system{}, a domain-specific LLM trained through a three-stage post-training pipeline that enhances SFC capabilities while preserving general performance. Extensive evaluations confirm that \system{} provides a reliable, high-performance solution for safety-sensitive fitness applications. 
We believe our approach can advance the real-world deployment of LLMs in fitness coaching and will inspire future research in developing specialized models in other vertical domains.

\bibliographystyle{IEEEtran}
\bibliography{cite}

\end{document}